\documentclass{article}

\usepackage{arxiv}

\usepackage[utf8]{inputenc} 
\usepackage[T1]{fontenc}    
\usepackage{hyperref}       
\usepackage{url}            
\usepackage{booktabs}       
\usepackage{amsfonts}       
\usepackage{nicefrac}       
\usepackage{microtype}      
\usepackage{graphicx}       
\usepackage{amsmath}
\usepackage{algorithm}
\usepackage{subcaption}
\usepackage{multirow}
\usepackage{array}

\title{JGAN: A Joint Formulation of GAN for Synthesizing Images and Labels}

\author{
  Minje Park \\
  Intel Corporation\\
  \texttt{minje.park@intel.com} \\
}

\begin{document}
\maketitle

\begin{abstract}
Image generation with explicit condition or label generally works better than unconditional methods. In modern GAN frameworks, both generator and discriminator are formulated to model the conditional distribution of images given with labels. In this paper, we provide an alternative formulation of GAN which models the joint distribution of images and labels. There are two advantages in this joint formulation over conditional approaches. The first advantage is that the joint formulation is more robust to label noises if it's properly modeled. This alleviates the burden of making noise-free labels and allows the use of weakly-supervised labels in image generation. The second is that we can use any kinds of weak labels or image features that have correlations with the original image data to enhance unconditional image generation. We will show the effectiveness of our joint formulation on CIFAR10, CIFAR100, and STL dataset with the state-of-the-art GAN architecture.
\end{abstract}

\keywords{Machine Learning \and Computer Vision and Pattern Recognition \and Generative Adversarial Networks \and Image Synthesis}

\section{Introduction}\label{sec:intro}

Due to the success of Generative Adversarial Network (GAN) for modeling distributions of real world data, it has been widely used for image generation. After the first introduction from Goodfellow and his colleagues~\cite{Goodfellow:2014}, many researchers have improved its stability and accuracy by adopting new loss functions~\cite{Arjovsky:2017, Berthelot:2017}, designing new network architectures~\cite{Salimans:2016, Gulrajani:2017}, improving training process and regularization~\cite{Gulrajani:2017, Miyato:2018spectral}, imposing conditions~\cite{Mirza:2014, Denton:2015, Reed:2016, Dumoulin:2017, Dumoulin:2017b, deVries:2017, Miyato:2018cgan}, and inventing progressive methods~\cite{Karras:2018}. Among them imposing explicit conditions is one of the easiest ways of improving the quality of image generation if there exist well-defined labels. In modern GAN frameworks, both generator and discriminator are formulated to model the conditional distribution of images given with labels.

\begin{figure*}[t]
    \centering
    \includegraphics[scale=0.35]{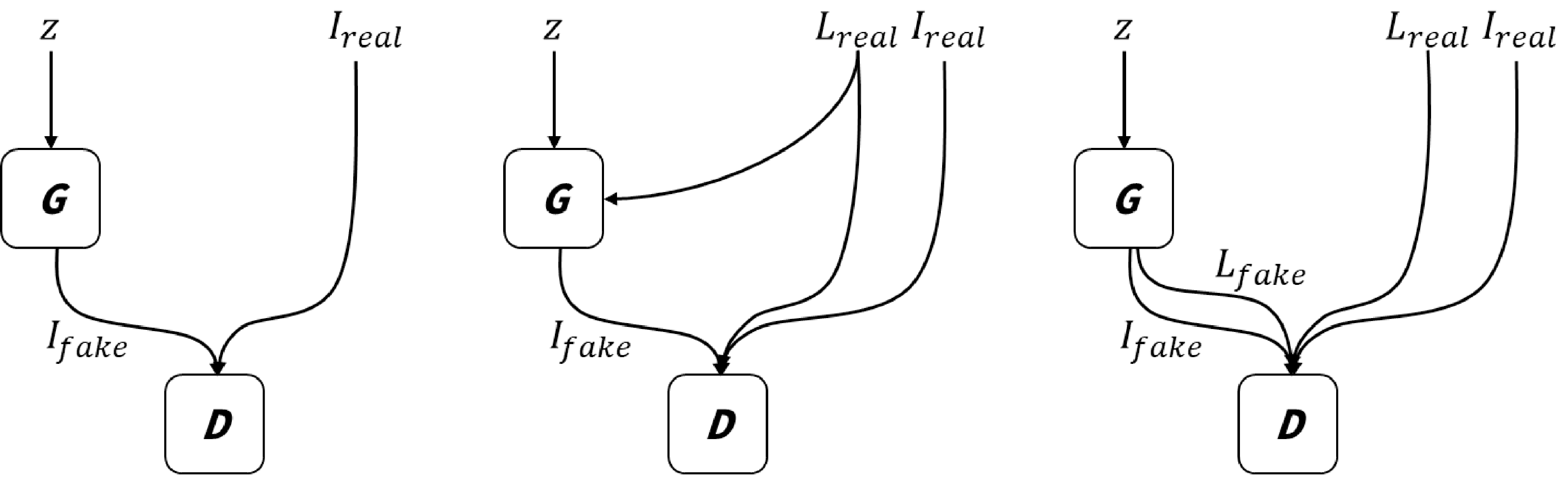}
    \caption{Three different GAN formulations: (left) Unsupervised GAN modeling $p(I|z)$; (middle) Conditional GAN modeling $p(I|z, L)$; (right, ours) Joint formulation of GAN modeling $p(I, L|z)$, generating images $I_{fake}$ and labels $L_{fake}$ simultaneously.}\label{im:intro}
\end{figure*}

In this paper, we propose an alternative formulation of GAN which models the joint distribution of images and labels. We will show that there are two advantages of this joint formulation over conditional approaches. The first advantage is that the joint formulation is more robust to label noises if it's properly modeled. Typical labels used in image synthesis are annotated by human workers or generated by other machine learning methods. It is generally difficult to guarantee the completeness or correctness of labels for large-scale data. Since conditional image generation regards labels as a given constraint or strong hypothesis, noises in labels may degenerate the quality of image generation. Our joint formulation regards labels as a piece of additional information to model the joint distribution. It could be more robust to the noises in labels since the joint probability distribution assumes no strong conditional dependence between images and labels. We will show the joint formulation provides the same level of image generation quality with defect-free labels and becomes more robust to noises in labels. Second and more importantly, we can use any kind of weak labels or additional information which have correlation with the original image data to enhance unconditional image generation since our joint GAN formulation doesn't require those labels when generating images but actually generates them along with images. In a conventional conditional formulation, it's impossible to feed these additional data into the generator since we don't know what kind of data should be added to the generator. Our experiment shows better image generation is possible without feeding labels or those additional information explicitly. Our contribution is summarized as follows:

\begin{itemize}
\item We propose a novel GAN formulation that models the joint distribution of images and labels, and show that this joint formulation increases the robustness on noisy or weak labels.
\item We demonstrate that this joint formulation can be used to increase the quality of unconditional image generation by incorporating weak labels or additional information which have correlation with the original image data into training process. Since the labels or those additional information are used only for training and our GAN generates both images and labels, we don't need to feed labels when generating images.
\end{itemize}

\section{A Joint formulation of GAN for modeling \texorpdfstring{$p(\mathbf{I, L})$}{p(I, L)}}\label{sec:jgan}

The standard adversarial loss for the discriminator $D$ for modeling the conditional probability $p(\mathbf{I}|\mathbf{L})$, in which $\mathbf{I}$ and $\mathbf{L}$ are images and labels respectively, is given by:
\begin{equation}
\begin{split}
    l(D)= & -E_{q(\mathbf{L})}[E_{q(\mathbf{I|L})}[log(D(\mathbf{I,L})]] \\ 
          & -E_{p(\mathbf{L})}[E_{p(G_{\mathbf{I}}(z)|\mathbf{L})}[log(1-D(G_{\mathbf{I}}(z), \mathbf{L}))]]
\end{split}
\end{equation}
, where $z$ is input noise, and $q$ and $p$ are the true distribution and the generator distribution, respectively. The generator loss is defined as:
\begin{equation}
    l(G)=-E_{p(\mathbf{L})}[E_{p(G_{\mathbf{I}}(z)|\mathbf{L})}[log(D(G_{\mathbf{I}}(z), \mathbf{L})]].
\end{equation}

In our joint formulation, we rewrite the discriminator and generator losses with a new generator $G_{\mathbf{I}, \mathbf{L}}(z)$, which  generate both $\mathbf{I}$ and $\mathbf{L}$ jointly, as follows:
\begin{equation}\label{eq:jgan}
\begin{split}
    l(D)= & -E_{q(\mathbf{L})}[E_{q(\mathbf{I|L})}[log(D(\mathbf{I}, \mathbf{L})]] \\
          & -E_{p(G_{\mathbf{I}, \mathbf{L}}(z))}[log(1-D(G_{\mathbf{I,L}}(z)))],
\end{split}
\end{equation}
\begin{equation}
    l(G)=-E_{p(G_{\mathbf{I,L}}(z))}[log(D(G_{\mathbf{I,L}}(z)))].
\end{equation}

As you can see, no modification is made on the discriminator since the discriminator has already a joint formulation which takes $p(\mathbf{L})$ and $p(\mathbf{I|L})$ (with the assumption of conditional dependence), and $G_{\mathbf{I, L}}$ generates $\mathbf{I}$ and $\mathbf{L}$, simultaneously. Figure~\ref{im:intro} illustrates the basic difference between the conditional and our joint formulation of exploiting labels.

Benefits of joint formulation over conditional formulation are limited when there exist well-defined labels, which are made carefully by human workers or  external oracles. It's well-known that modeling joint distribution is generally a more difficult task than modeling conditional distribution due to its increased dimension in probability distribution. Thus the discriminator represents the joint distribution by  the lower dimension probability distributions $p(\mathbf{L})$ and $p(\mathbf{I|L})$. The only difference here is how we can incorporate the label in generators. Common choices of imposing condition on generators are input or hidden concatenation~\cite{Mirza:2014, Denton:2015, Reed:2016, Dumoulin:2017} and conditional batch normalization~\cite{Dumoulin:2017b, deVries:2017}. Our joint formulation doesn't require labels as a condition but actually generates labels with the given input noise along with images. To do this we add an additional function approximator as a part of the generator (refer to the Experiment section for the choices of the label function approximators). Since this joint formulation doesn't use labels as a prior for lowering the dimension of the probability distribution of the data, it can be more robust to the noises in labels if we can properly model the joint distribution in the original dimension of probability distribution. A possible drawback of this joint approach over traditional condition GAN is that we lose the controllability of image generation. However, for some scenarios, if there exist several labels, both our and conditional GAN can be used together to achieve the controllability on clean labels and the robustness on noisy labels, possibly not generated by oracle but automatically generated by other system. We will explain this in details at the next section.

\subsection{Boosting Unsupervised Image Generation}\label{sec:unsuper}

\begin{figure}[t!]
\centering\includegraphics[scale=0.25]{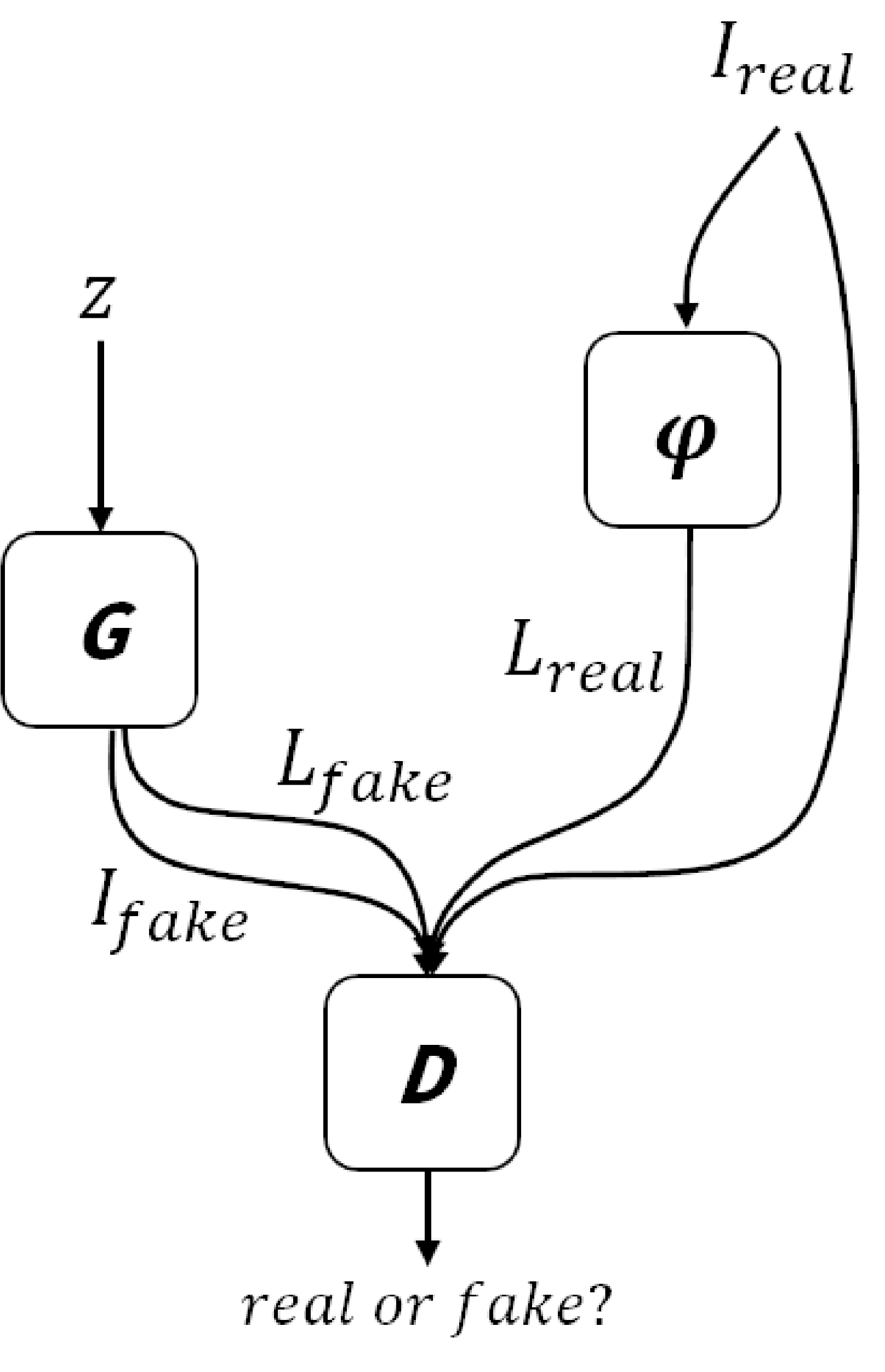}
\caption{Enhancing unsupervised image generation by using an additional label predictor $\varphi$, which generates weak (or pseudo) label $L_{real}$. The label generator part of JGAN captures the distribution of $\varphi$. $L_{real}$ and the generated label $L_{fake}$ are fed into the discriminator in a conventional way during training.}\label{im:boost}
\end{figure}

With our joint formulation we can add additional information, which has dependence on the original data, as a weak label for the generator. Figure~\ref{im:boost} illustrates how we can add output from other classification network $\varphi$ for boosting the quality of unsupervised image generation. Typical choices for $\varphi$ are class prediction results from other tasks like ImageNet classification and object detection. You can also use  unsupervised learning algorithms like $k$-means clustering or autoencoders~\cite{Hinton:2006}. This is an unique advantage of JGAN over conditional GANs since the additional information is modeled simultaneously by the generator, and the discriminator uses this fake information as a condition for the decision. As you can see in Equation~\ref{eq:jgan}, the discriminator actually models the joint distribution with the prior equals to the training label distribution. This additional information can boost the quality of synthesized images since it can act like weak labels for the discriminator. In conventional conditional GANs, this is practically impossible since it's hard to feed $\varphi(I_{fake})$ while generating images.

\section{Experiment}\label{sec:exp}

We used CIFAR10, CIFAR100, and STL for our experiment, and resized STL images to 48x48 from its original size of 96x96. For all experiments, we fixed the discriminator architecture for assessing the effect of our joint formulation. We followed the design used by Miyato et al.~\cite{Miyato:2018spectral} as a baseline framework for the entire experiment. Table~\ref{tb:gen} and~\ref{tb:disc} show the architecture of our generator and discriminator respectively. We removed batch normalization and applied spectral normalization to all layers of the discriminator. We used one discriminator update for each generator update, and all results are evaluated at 100K generator updates except STL case, in which we used 200K generator updates for better convergence. We used 0.0004 for the learning rate for the discriminator, and 0.0001 for the generator with Adam optimizer with $\beta_{1}=0.5$ and $\beta_{2}=0.999$. We reported the average inception score~\cite{Salimans:2016} of the last five epochs of several runs rather than the best score~\cite{Lucic:2018}.

\begin{figure*}[h]
\centering\includegraphics[scale=0.95]{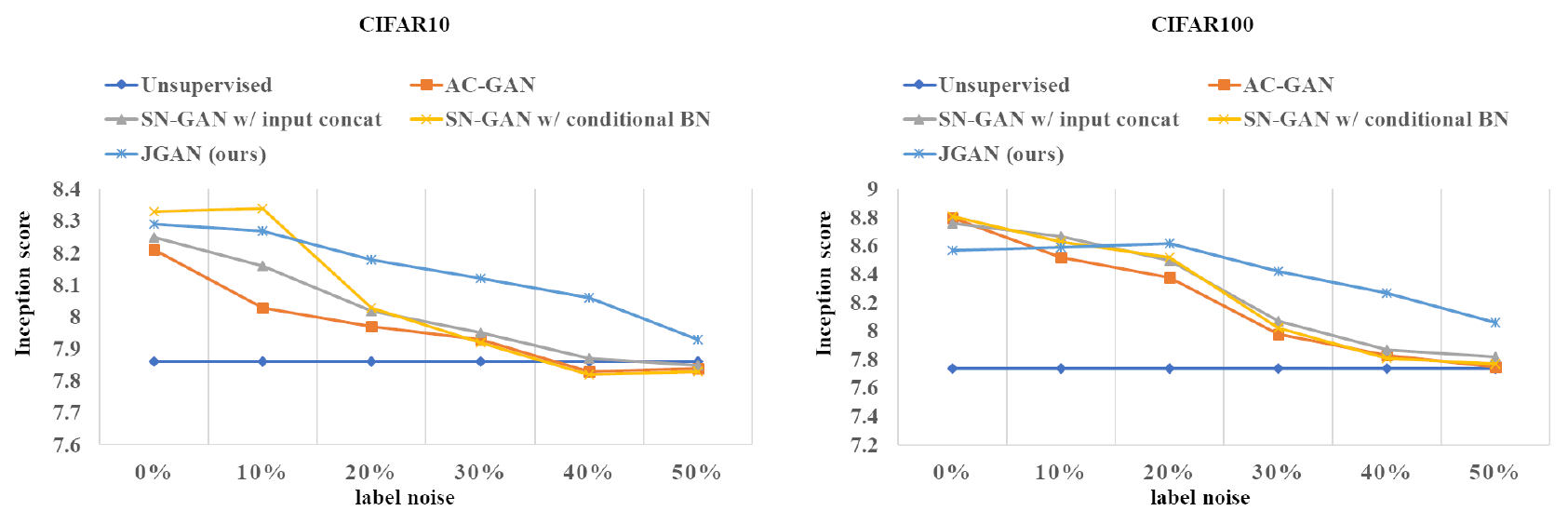}
\caption{Inception scores on CIFAR10 and CIFAR100 with different label noise ratios. Graphical illustration of Table~\ref{tb:cifar10} (left) and Table~\ref{tb:cifar100} (right). JGAN shows robustness on label noise in both cases.}\label{im:noise_graph}
\end{figure*}

We first show our joint formulation is as good as the conditional formulation when modeling the conditional distribution $p(\mathbf{I, L})$ for clean labels, and more robust to label noises. We used input concatenation~\cite{Reed:2016, Dumoulin:2017} and conditional batch normalization~\cite{Dumoulin:2017b, Miyato:2018cgan} for the generator for comparison, and the projection discriminator with hinge loss proposed by Miyato et al.~\cite{Miyato:2018cgan}, which shows the state-of-the-art result for conditional image generation. To generate labels, we added a function approximator composed of several neural network layers right after the last ReLU layer of the generator in Table~\ref{tb:gen}. Table~\ref{tb:labelgen} describes the network architecture for the label generation part of the generator. Dropout~\cite{Srivastava:2014} is applied to all dense layers of the label generator with the rate of 0.5 to avoid overfitting. We added label noises by randomly selecting a subset of the entire dataset and then applied a random offset for each selected label. Table~\ref{tb:noise} and Figure~\ref{im:noise_graph} summarize the result of inception score changes according to the amount of label noises. As you can see, our joint formulation shows a competitive result on clean labels, and remains robust even on high label noise ratios.

\begin{table*}[htbp]
    \caption{Inception scores on CIFAR10 and CIFAR100 with different label noise ratios. Note that the joint formulation is more robust than the conditional one at high noise ratios. The conditional formulation have almost no benefit over 40\% label noises but the joint formulation has improvements (7.86 vs 8.06 (ours) and 7.86 vs 7.93 (ours)). This improvement becomes more noticeable in CIFAR100 case (7.74 vs 8.27 (ours) and 7.74 vs 8.06 (ours)).}
    \label{tb:noise}
    \centering
    \begin{subtable}[h]{0.9\textwidth}
        \centering
        \begin{tabular}{c c c c c c c}
        \hline
         & \multicolumn{6}{c}{Label noise ratio} \\ \hline
        Method & 0\% & 10\% & 20\% & 30\% & 40\% & 50\% \\ \hline
        Unsupervised~\cite{Miyato:2018spectral} & 7.86 & 7.86 & 7.86 & 7.86 & 7.86 & 7.86 \\
        AC-GAN~\cite{Odena:2017} & 8.21 & 8.03 & 7.97 & 7.93 & 7.83 & 7.84 \\
        SN-GAN w/ input concatenation~\cite{Miyato:2018spectral} & 8.25 & 8.16 & 8.02 & 7.95 & 7.87 & 7.85 \\
        SN-GAN w/ conditional batchnorm~\cite{Miyato:2018spectral} & \textbf{8.33} & \textbf{8.34} & 8.03 & 7.92 & 7.82 & 7.83 \\
        Joint (ours) & 8.29 & 8.27 & \textbf{8.18} & \textbf{8.12} & \textbf{8.06} & \textbf{7.93} \\
        \hline
        \end{tabular}
        \caption{CIFAR10}\label{tb:cifar10}
    \end{subtable}

    \begin{subtable}[h]{0.9\textwidth}
        \centering
        \begin{tabular}{c c c c c c c}
        \hline
         & \multicolumn{6}{c}{Label noise ratio} \\ \hline
        Method & 0\% & 10\% & 20\% & 30\% & 40\% & 50\% \\ \hline
        Unsupervised~\cite{Miyato:2018spectral} & 7.74 & 7.74 & 7.74 & 7.74 & 7.74 & 7.74 \\
        AC-GAN~\cite{Odena:2017} & 8.80 & 8.52 & 8.38 & 7.98 & 7.83 & 7.75 \\
        SN-GAN w/ input concatenation~\cite{Miyato:2018spectral} & 8.76 & \textbf{8.67} & 8.50 & 8.07 & 7.87 & 7.82 \\
        SN-GAN w/ conditional batchnorm~\cite{Miyato:2018spectral} & \textbf{8.81} & 8.63 & 8.52 & 8.02 & 7.81 & 7.77 \\
        Joint (ours) & 8.57 & 8.59 & \textbf{8.62} & \textbf{8.42} & \textbf{8.27} & \textbf{8.06} \\
        \hline
        \end{tabular}
        \caption{CIFAR100}\label{tb:cifar100}
    \end{subtable}
\end{table*}

Our next experiment is focused on improving unconditional image generation by incorporating an additional information. We used the class probability of inception network as a staring point of this addtional information. We used the same inception network version used in~\cite{Salimans:2016}. Since it has a probability distribution of 1000 classes and it's difficult to find the optimal network architecture to capture this high dimensional probability distribution, we applied truncated singular value decomposition (SVD) to reduce its dimension to 64 to simplify the problem. We applied softmax to the output of truncated SVD to make it a probability distribution in lower dimensional space. Table~\ref{tb:weak} summarizes the comparison result between unsupervised and and joint image generation. We used the same network architecture for both unsupervised and joint settings except additional label function approximation. We used a label generator slightly different from the ones used in Table~\ref{tb:labelgen}. Table~\ref{tb:weakgen} describes the network for weak label generation. As you can see, JGAN consistently generates images with higher inception and FID~\cite{Heusel:2017} scores compared to unsupervised ones. We have achieved the best unsupervised image generation score in STL dataset compared to \cite{Gulrajani:2017} and \cite{Grinblat:2017}, that reported 9.05 and 9.50, respectively. Note that our baseline implementation achieved a better result due to different network architecture and training process but our joint formulation achieved even higher inception and FID scores.

\begin{table}[htbp]
\caption{Inception scores and FIDs with unsupervised image generation on CIFAR10, CIFAR100, and STL. Weak labels from Inception pool3 are used in training JGAN. Note that accuracy numbers of reference implementation differ slightly due to our reimplementation with architectural change and training hyperparameters}\label{tb:weak}
\centering
\begin{tabular}{l l l}
\hline
Dataset and method & IS & FID\\
\hline
CIFAR10 & & \\
\hspace{0.5cm}Real & 11.2 & 7.62 \\
\hspace{0.5cm}WGAN-GP~\cite{Gulrajani:2017} & 7.86 & \\
\hspace{0.5cm}SN-GAN~\cite{Miyato:2018spectral} & 7.86 & 24.5 \\
\hspace{0.5cm}JGAN (ours) & \textbf{7.99} & \textbf{22.7} \\
\hline
CIFAR100 & & \\
\hspace{0.5cm}Real & 14.17 & 8.93 \\
\hspace{0.5cm}WGAN-GP~\cite{Gulrajani:2017} & 7.68 & \\
\hspace{0.5cm}SN-GAN~\cite{Miyato:2018spectral} & 7.74 & 32.8 \\
\hspace{0.5cm}JGAN (ours) & \textbf{8.02} & \textbf{30.6} \\
\hline
STL & & \\
\hspace{0.5cm}Real & 26.6 & 7.92 \\
\hspace{0.5cm}WGAN-GP~\cite{Gulrajani:2017} & 9.05 & \\
\hspace{0.5cm}SN-GAN~\cite{Miyato:2018spectral} & 9.73 & 35.2 \\
\hspace{0.5cm}Class splitting GAN~\cite{Grinblat:2017} & 9.50 & \\
\hspace{0.5cm}JGAN (ours) & \textbf{10.08} & \textbf{31.7} \\
\hline
\end{tabular}
\end{table}

\begin{figure*}[t!]
\centering\includegraphics[scale=0.61]{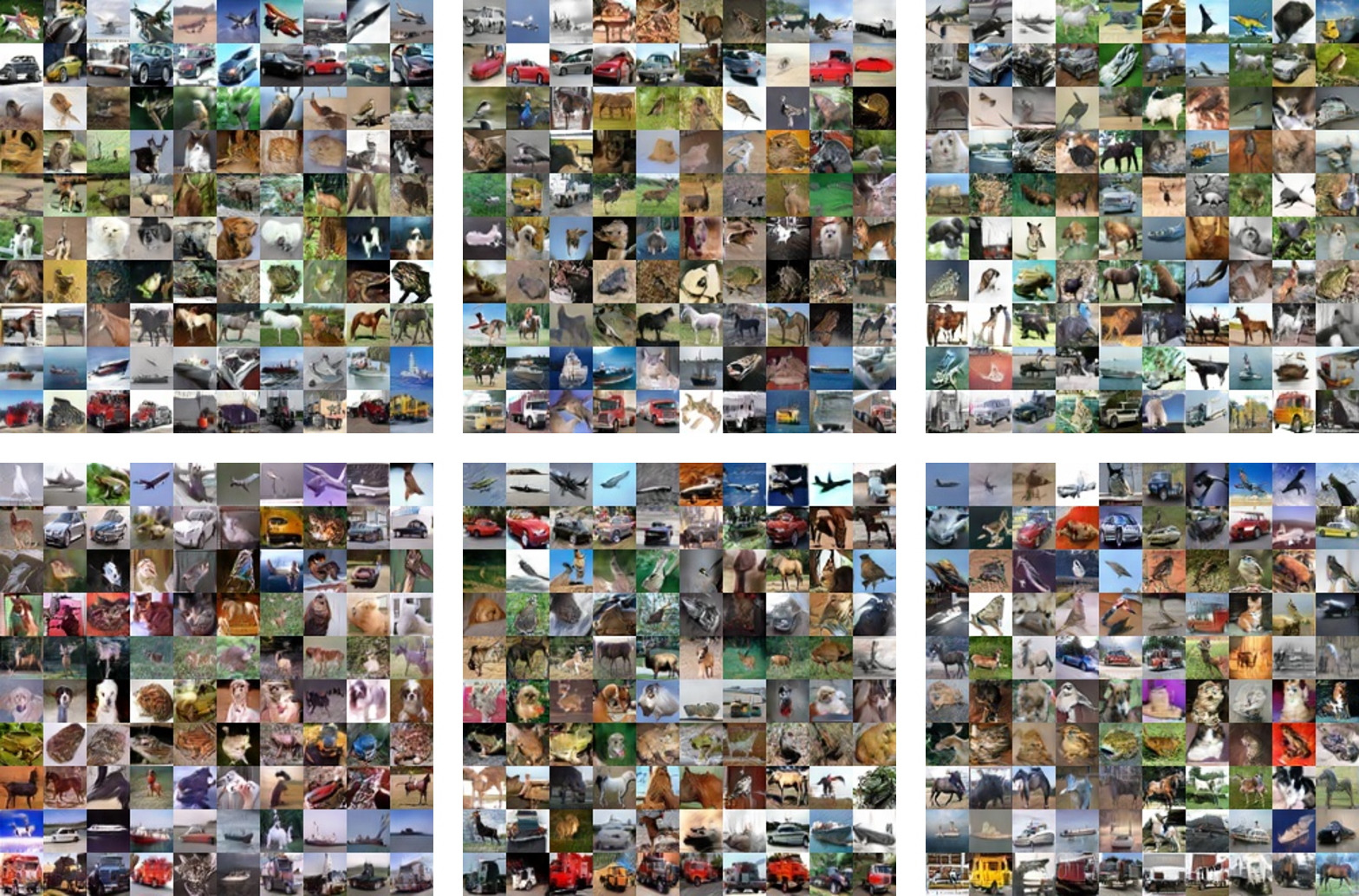}
\caption{Comparison of images generated by (top row) conditional GAN and (bottom row) joint GAN on CIFAR10 with noisy labels. (from left to right) Generated images with clean label (0\% noise), 20\% noise, and 40\% noise. In each sub-figure, rows are class ids and columns are random samples of each class id. Our JGAN shows a better inception score than conditional GAN (Refer to table~\ref{tb:cifar10} for inception score of each case).}\label{im:noise_result}
\end{figure*}

\begin{figure*}[t!]
\centering\includegraphics[scale=0.51]{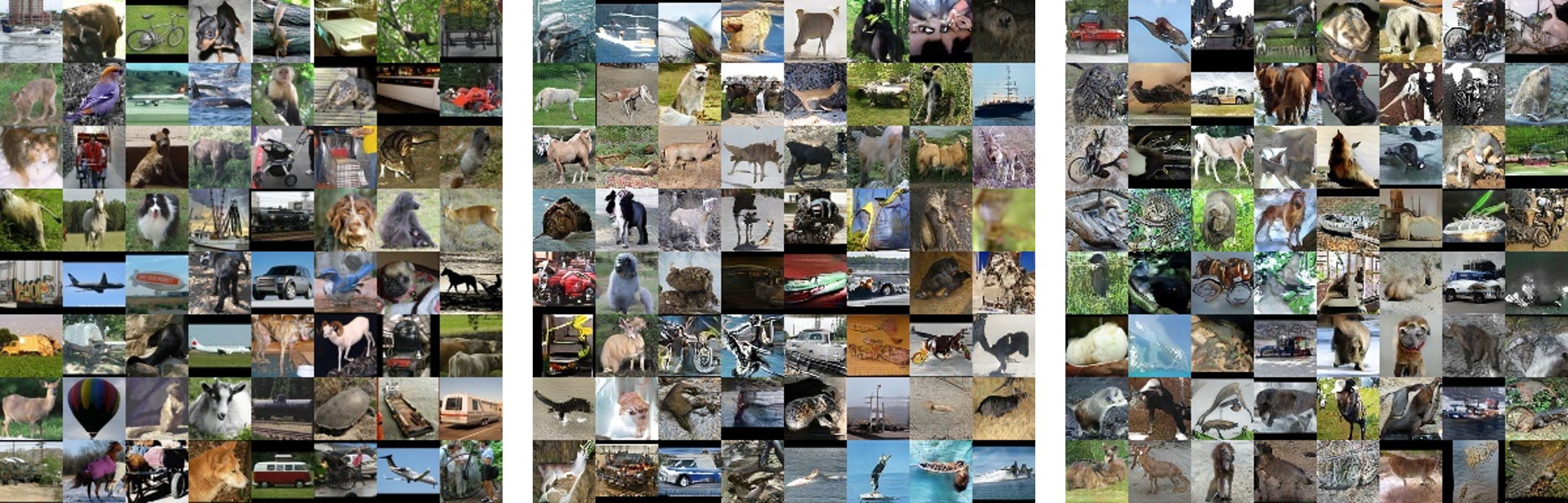}
\caption{Comparison of (left) real unlabeled STL dataset, (middle) images generated by  unsupervised GAN, and (right) joint GAN with weak label from ImageNet classification task, which shows a better inception score than unsupervised image synthesis (refer to table~\ref{tb:weak}).}\label{im:stl_result}
\end{figure*}

\section{Conclusion}\label{sec:conc}
In this paper, we propose a novel GAN framework that models the joint probabilistic distribution of images and labels. We showed that this joint formulation can generate as good image quality as the conventional conditional image generation with clean labels, and remains robust when there exist noises in labels. We also applied our method to improve the image quality of unconditional image generation by incorporating additional information which has correlation with the original image data. We think this joint formulation can provide an easy way to feed many kinds of relevant information or weak labels into the GAN framework with a simple modification of the generator. There are several interesting future works like finding optimal network architectures for the label generator and testing with other methods for generating additional information we can use with our joint formulation. Even though we used images as our main target domain, we expect our formulation works for other domains as well.

\begin{table}[ht]
    \caption{Generator, $D_{b}=4$ for CIFAR10 and CIFAR100, and $D_{b}=6$ for STL}\label{tb:gen}
    \centering
    \begin{tabular}{l}
    \hline\hline
    $z\in\mathbb{R}^{128}\sim\mathcal{N}(0, \mathbf{I})$ \\ \hline
    dense, $D_{b} \times D_{b} \times 256$ \\ \hline
    ResBlock $3\times3$, upscale, 256 \\ \hline
    ResBlock $3\times3$, upscale, 256 \\ \hline
    ResBlock $3\times3$, upscale, 256 \\ \hline
    BN, ReLU, $3\times3$ conv, 3, tanh \\ \hline
    \hline
    \end{tabular}
\end{table}
    
\begin{table}[ht]
    \caption{Discriminator, $D_{f}=32$ for CIFAR10 and CIFAR100, and $D_{f}=48$ for STL. Spectral normalization is applied to all layers. For conditional image generation, we used projection disciminator proposed by Miyato et al.~\cite{Miyato:2018cgan}.}\label{tb:disc}
    \centering
    \begin{tabular}{l}
    \hline\hline
    RGB image $\in\mathbb{R}^{D_{f} \times D_{f} \times 3}$ \\ \hline
    ResBlock, $3\times3$, downscale, 128 \\ \hline
    ResBlock, $3\times3$, downscale, 128 \\ \hline
    ResBlock, $3\times3$, same, 128 \\ \hline
    ResBlock, $3\times3$, same, 128 \\ \hline
    ReLU, Global Sum Pooling \\ \hline
    dense, 1 \\ \hline
    \hline
    \end{tabular}
\end{table}

\begin{table}[ht]
    \caption{Label generation part of the generator, $D_{r}=32$ for CIFAR and $D_{r}=48$ for STL, $C_{l}=128$ for CIFAR10 and STL and $C_{l}=256$ for CIFAR100, $D_{o}=10$ for CIFAR10 and STL and $D_{o}=100$ for CIFAR100. We used one-hot vector representation for labels.}
    \label{tb:labelgen}
    \centering
    \begin{tabular}{c}
    \hline\hline
    Output of the last ReLU of the generator $\in\mathbb{R}^{D_{r} \times D_{r} \times 256}$  \\ \hline
    $7 \times 7$ conv, stride=4, 256 \\ \hline
    BN, ReLU, dense, $C_{l}$ \\ \hline
    BN, ReLU, dense, $C_{l}$ \\ \hline
    BN, ReLU, dense, $D_{o}$ \\ \hline
    \hline
    \end{tabular}
\end{table}

\begin{table}[ht]
    \caption{Label generation part of the generator, $D_{r}=32$ for CIFAR and $D_{r}=48$ for STL, $C_{l}=128$ and $D_{o}=64$ for all cases.}
    \label{tb:weakgen}
    \centering
    \begin{tabular}{c}
    \hline\hline
    Output of the last ReLU of the generator $\in\mathbb{R}^{D_{r} \times D_{r} \times 256}$  \\ \hline
    $7 \times 7$ conv, stride=4, 256 \\ \hline
    BN, ReLU, dense, $C_{l}$ \\ \hline
    BN, ReLU, dense, $C_{l}$ \\ \hline
    BN, ReLU, dense, $C_{l}$ \\ \hline
    BN, ReLU, dense, $D_{o}$ \\ \hline
    \hline
    \end{tabular}
\end{table}

\bibliographystyle{unsrt}  
\bibliography{references}

\end{document}